\documentclass[conference, onecolumn]{./template/IEEEtran}
\IEEEoverridecommandlockouts
\usepackage{cite}
\usepackage{amsmath,amssymb,amsfonts}
\usepackage{algorithmic}
\usepackage{graphicx}
\usepackage{textcomp}
\usepackage{xcolor}
\def\BibTeX{{\rm B\kern-.05em{\sc i\kern-.025em b}\kern-.08em
    T\kern-.1667em\lower.7ex\hbox{E}\kern-.125emX}}

\usepackage{hyperref}
\usepackage{subcaption}
\usepackage{booktabs}
\usepackage{wrapfig}

\usepackage{orcidlink}

\newcommand{\fows}{\texttt{FOWS}}
\newcommand{\gotcha}{\texttt{GOTCHA}}

\newcommand{\orF}{OR$_{\texttt{F}}$}
\newcommand{\sims}{SS$_{\texttt{F}}$}
\newcommand{\gh}{GH$_{\texttt{F}}$}
\newcommand{\fd}{FD$_{\texttt{F}}$}
\newcommand{\orG}{OR$_{\texttt{G}}$}
\newcommand{\dfl}{DFL$_{\texttt{G}}$}
\newcommand{\fsgan}{FSGAN$_{\texttt{G}}$}

\newcommand{\wocc}{\textbf{occ}}
\newcommand{\woocc}{\textbf{no-occ}}

\newcommand{\mnet}{\textit{MobNet}}
\newcommand{\enet}{\textit{EffNetB4}}
\newcommand{\xnet}{\textit{Xception}}
\newcommand{\enetPRE}{\textit{ICPR2020}}
\newcommand{\xnetPRE}{\textit{NeurIPS2023}}

\newcommand{\ba}{B-ACC}
\newcommand{\acc}[1]{ACC(#1)}
\newcommand{\auc}{AUC}
\newcommand{\eer}{EER}

\usepackage{tikz}

\newcommand\submittedtext{%
  \footnotesize This work has been submitted to the IEEE for possible publication. Copyright may be transferred without notice, after which this version may no longer be accessible.}

\newcommand\submittednotice{%
\begin{tikzpicture}[remember picture,overlay]
\node[anchor=south,yshift=10pt] at (current page.south) {\fbox{\parbox{\dimexpr0.65\textwidth-\fboxsep-\fboxrule\relax}{\submittedtext}}};
\end{tikzpicture}%
}

\renewcommand\fbox{\fcolorbox{red}{white}}
\begin{document}

%% --------------------------------------------------------------------- %%
%% possible title: Investigating challenge-based authentication in RIDP  %%
%% --------------------------------------------------------------------- %%

\title{Spotting tell-tale visual artifacts in face swapping videos: strengths and pitfalls of CNN detectors\\
}

% \author{\IEEEauthorblockN{Riccardo Ziglio} 
% \IEEEauthorblockA{\textit{--} \\
% \textit{--}\\
% -- \\
% --}
% \and
% \IEEEauthorblockN{ Cecilia Pasquini}
% \and
% \IEEEauthorblockN{ Silvio Ranise}

%% -------------------------------------------------------------------------- %%
%% works - missing orcid and mail %%
\author{
\IEEEauthorblockN{Riccardo Ziglio\textsuperscript{1,3}, % \orcidlink{0009-0000-4373-3894}, 
Cecilia Pasquini\textsuperscript{1}, % \orcidlink{0000-0002-2125-6983}, 
Silvio Ranise\textsuperscript{1,2} % \orcidlink{0000-0001-7269-9285}
}
\IEEEauthorblockA{\textsuperscript{1}\textit{Center for Cybersecurity, Fondazione Bruno Kessler, Trento, Italy}\\
\textsuperscript{2}\textit{Department of Mathematics, University of Trento, Italy}\\
\textsuperscript{3}\textit{DIBRIS, University of Genoa, Italy} \\
% \textit{Department of Informatics, Bioengineering, Robotics and System Engineering, University of Genoa, Italy} \\
{\{rziglio, c.pasquini, ranise\}}@fbk.eu
% email address or ORCID
%% close IEEEauthorblock
}
%% close author
}
%% -------------------------------------------------------------------------- %%

\maketitle

%% -------------------------------------------------------------------------- %%
%% Add submitted notice
\submittednotice
%% -------------------------------------------------------------------------- %%

\begin{abstract}

Face swapping manipulations in video streams represents an increasing threat in remote video communications, due to advances in automated and real-time tools.
Recent literature proposes to characterize and exploit visual artifacts introduced in video frames by swapping algorithms when dealing with challenging physical scenes, such as face occlusions.  
This paper investigates the effectiveness of this approach by benchmarking CNN-based data-driven models on two data corpora (including a newly collected one) and analyzing generalization capabilities with respect to different acquisition sources and swapping algorithms. The results confirm excellent performance of general-purpose CNN architectures when operating within the same data source, but a significant difficulty in robustly characterizing  occlusion-based visual cues across datasets. This highlights the need for specialized detection strategies to deal with such artifacts. 
\end{abstract}

\begin{IEEEkeywords}
face swapping, face verification, remote video calls, forensic detection
\end{IEEEkeywords}

%% -------------------------------------------------- %%

\section{Introduction}

% \cite{IWBFpaper}

% test citations: \cite{cyborgLoss}, \cite{dang2020detectiondigitalfacemanipulation}, \cite{OC-FakeDect}, \cite{iProof_ThreatIntelligenceReport}, \cite{handbookInjectionAttacks}
% \cite{iProof_ThreatIntelligenceReport}

The synthesis and manipulation of facial images and videos have achieved increasingly hyper--realistic results in recent years \cite{age-synthetic-reality}, leading to numerous research efforts for the automated identification of non-genuine visual data \cite{face-manipulation-and-fake-detection} \cite{media-forensics-and-df}. 

The wide majority of the detection techniques proposed in the literature operates in a passive fashion (i.e., assuming no or little a priori information on the media generation and distribution pipeline) \cite{df2024}. Data-driven approaches based on deep networks are predominant, with an extensive use of general-purpose architectures originally devised for image classification, revisited for these tasks \cite{MilanPaper}\cite{ff++} \cite{DFB}.
Practical applications range from the analysis of user-generated content on the web for fact checking purposes to the validation of digital visual evidence in forensic investigations.

A highly relevant scenario where detecting manipulations becomes essential is remote live video communications, where one (or more) subject stands in front of a camera capturing the video scene in real-time and interacts remotely with another subject or with an automated interface. 
If not properly protected, such scenario might be vulnerable to advanced impersonation attacks enabled by real-time video face manipulation tools,  potentially leading to severe security issues.
% It is the case of video calls, which 
It is the case of video calls, which recently emerged as a channel for perpetrating scams and frauds\footnote{\scriptsize \url{https://edition.cnn.com/2024/02/04/asia/deepfake-cfo-scam-hong-kong-intl-hnk/index.html}}. In fact, remote identity proofing processes (such as Know-Your-Customer applications) based on face verification \cite{ENISA_RemoteIDProofing}\cite{ ENISA_RemoteIDProofingA&C} \cite{DBLP:conf/pkdd/PasquiniPSR23} are threatened by these new tools
\cite{SeeingIsLiving} \cite{chapter2022}, in addition to known presentation attack vectors \cite{pres2017}. In this context, face swapping is a particularly powerful technique, as it allows an attacker to modify only the facial area (typically used for verifying the subject's identity), while leaving the rest of the scene under his control.

The detection of face swapping in remote video communications has so far been relatively little explored, with only a few approaches explicitly targeting this scenario \cite{farid,stamm}. Recently, the work in \cite{gotcha} has proposed to exploit the interactive nature of the setting to create a detection advantage, inspired by biometric challenge-response protocols \cite{cr}: the user performs (mostly voluntary) actions denoted as {\it challenges}, which are devised to interfere with manipulation algorithms and, in case of ongoing impersonation attack, produce visible visual artifacts in the video frame. While the quality levels of manipulation engines are rapidly improving, handling diversified challenge requests in real-time (with no post-processing options) represents a complex task. Face occlusions are a notable case where evident artifacts are likely to be produced, providing precious hints for human observers. 

This paves the way for automated detectors exploiting such tell-tale visual cues to identify swapped faces over genuine ones. The authors in \cite{gotcha} provide empirical evidence of a learning-based approach, validated on their proposed dataset.
However, a known crucial issue of data-driven solutions for manipulation detection is their ability to generalize in non-aligned testing modes \cite{media-forensics-and-df}\cite{df2024}, thus when testing data do not come from the very same data source as the training set. Thus, it remains an open question to which extent learned cues can effectively detect occlusion artifacts across data coming from different acquisition settings and manipulation algorithms.

The experimental analysis proposed in this paper investigates this scenario. We introduce a newly collected data corpus of genuine and manipulated videos systematically depicting subjects with face occlusions, as reported in Section \ref{dataset_collection}. This, together with the dataset in \cite{gotcha}, allows us to perform a cross-dataset analysis (described in Section \ref{sec:setup}) and benchmark the performance of CNN-based data-driven detection models in different experimental settings. The influence of both the data source and the frame content are analyzed, revealing insightful empirical evidence reported in Section \ref{sec:res}.

%% ------------------------------------ %%
\section{Dataset Collection} \label{dataset_collection}

We denote as {\fows} (Face Occlusion With Swapping) the newly collected dataset. Each video clip depicts a human subject in front of a camera who is asked to perform voluntary actions (also indicated as {\it challenges}) leading to face occlusions. 
In fact, in \cite{gotcha} those have proved highly effective in producing visual artifacts after face swapping and obtained a high usability score when rated by users
among the different considered alternatives (i.e., head movements, facial deformation, facial illumination). 
In particular, our subjects are asked to place in front of their face either their own hand ({\it hand occlusion} challenge) or a supplied rectangular object ({\it object occlusion} challenge) in predefined locations. 

\begin{figure}[t!]
% \centering
\begin{subfigure}{\linewidth}
\centering 
\includegraphics[width=\linewidth]{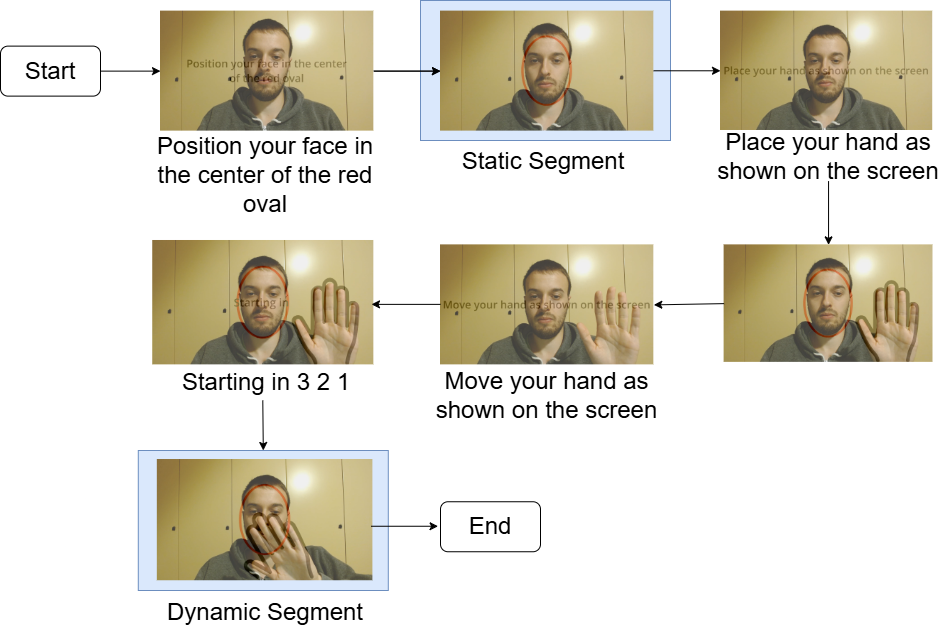}

(a)
\end{subfigure}
\hfill

\begin{subfigure}{\linewidth}
\centering

\includegraphics[width=0.22\linewidth]{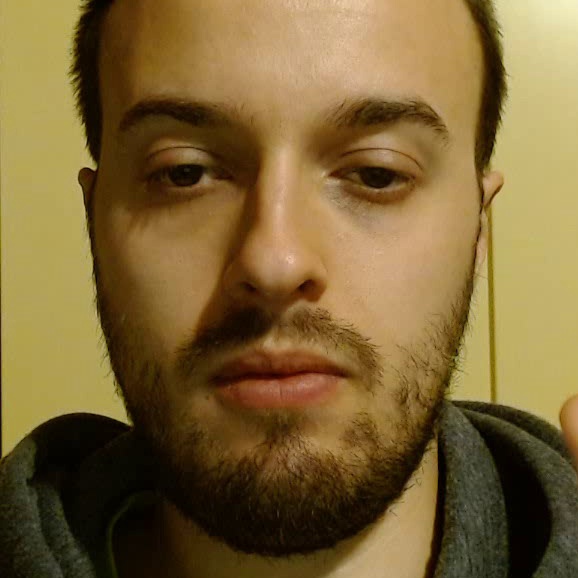}
\includegraphics[width=0.22\linewidth]{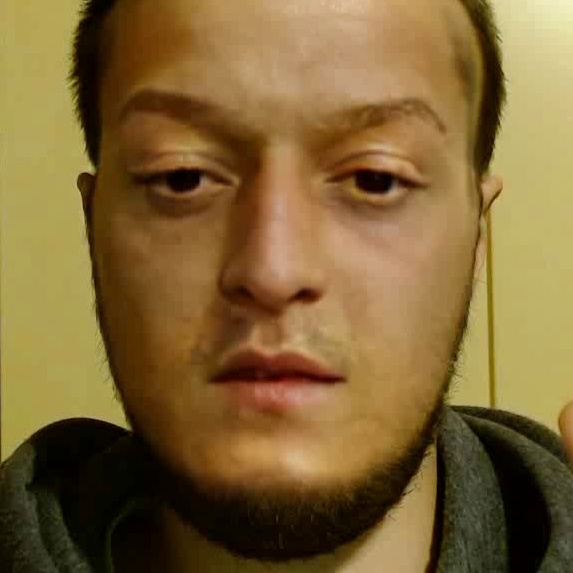}
\includegraphics[width=0.22\linewidth]{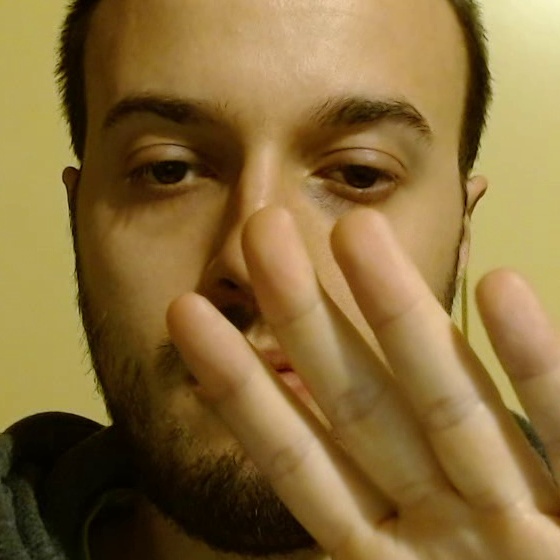}
\includegraphics[width=0.22\linewidth]{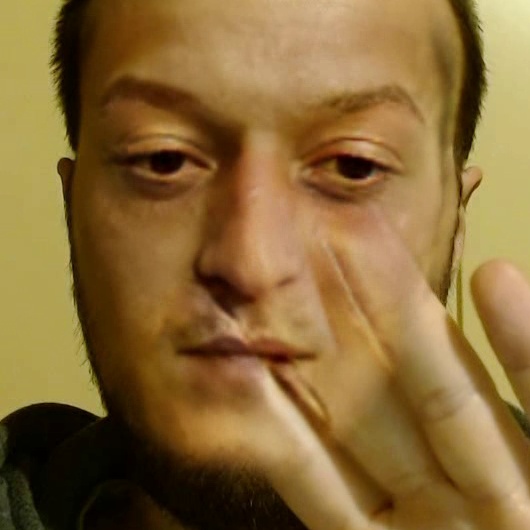}

(b)
\end{subfigure}

    \caption{(a) Pipeline of the video recording with the guiding videos for the hand occlusion challenge. (b) Examples of original (left) and manipulated (right) frames.
    }
    \label{fig:fows_ex}
\end{figure}

We developed a recording interface to streamline the video collection from volunteer users. The interface contains a short description and some visual examples of correct/incorrect positions for each of the challenges to be performed.
Then, during the actual execution, a {\it guiding video} displaying visual instructions on where to place the face and the occluding objects is overlaid on the current camera frame, as represented in Figure \ref{fig:fows_ex}(a). Subjects have been instructed to follow as closely as possible the motion shown in the guiding videos; for each clip, we visually verified the compliance of the user's motion with the displayed instructions at the recording phase.

As pointed out in \cite{gotcha}, multiple and diversified occlusions per user help in better detecting potential manipulation attacks. Therefore, we created three different variants of the hand and object occlusion challenges to be performed by each user, where the occluding objects follow different spatial trajectories and overlaps with diverse portions of the face.
Thus, six different guiding videos have been developed lasting around 40 seconds, all of them are structured as follows (see Figure \ref{fig:fows_ex}(a)): they start with a static temporal segment depicting only the subject's face in a prescribed location, followed by a more dynamic segment where the occluding object moves and overlaps the face (lasting around 15 seconds). 
We recorded videos from 7 volunteers, each of them recording the six challenge variants. All videos were captured with a Logitech C920 HD PRO webcam recording at 1080p resolution and 30 fps.
From the dynamic segment of each genuine video, three manipulated versions have been then generated through the recent face-swapping algorithms SimSwap (\sims) \cite{SimSwap_ChenCNG20}, GHOST (\gh) \cite{GHOST_Groshev} and FaceDancer (\fd) \cite{Facedancer_Rosberg}. These algorithms are particularly powerful as they perform face swapping from a single image of the target face. We used royalty-free pictures of celebrities as target faces. 
Examples of genuine and manipulated frames with and without face occlusion are reported in Figure \ref{fig:fows_ex}(b). In agreement with \cite{gotcha}, occlusion-based artifacts are present in all manipulated videos.

In total, the dataset is composed of 168 videos (42 genuine and 126 manipulated), and approximately 70k video frames.  
% 
% Examples of genuine and manipulated video frames are reported in Figure \ref{fig:fows_ex}(b) for the case of hand occlusion. In agreement with \cite{gotcha}, occlusion-based artifacts are present in all manipulated videos.

%% -------------------------------------------------- %%
 
%% ------------------------------------ %%

\section{Experimental Setup}
\label{sec:setup}

% This section describes the setup of our experimental analysis. The data and part of the code used in this study are available at \todo{insert link to project page FBK}.

The dataset and part of the code used used in this study are available at \url{https://st.fbk.eu/complementary/IWBF2025}.
% Part of the data and code used in this study will be made available upon acceptance.

\subsection{Datasets}
\label{ssec:data} 
% [original version of the sentence] 
Our experimental analysis involves the previously described {\fows} dataset and the one proposed in \cite{gotcha}, denoted as {\gotcha} dataset. Among the available datasets including face swapping manipulations, this is the only one where subjects systematically perform face occlusion and thus fits the purpose of our study. 
% [original portion of the sentence] this is the only one where subjects systematically perform face occlusion and thus fits the purpose of our study. 
The dataset involves 47 participants performing several challenges; it includes genuine videos and two corresponding swapped versions, obtained through the algorithms DeepFaceLab (\dfl) \cite{DFL} and FSGAN (\fsgan) \cite{FSGAN}. Data are provided as sequences of static frames.
In particular, we selected the data related to the hand and object occlusion challenges only; for each user, swapped videos from three different target faces for each challenge are considered among the available ones. 

Overall, from the union of the two datasets we can obtain seven data partitions: five corresponding to the swapping algorithms (denoted as {\sims}, {\gh}, {\fd}, {\dfl}, and {\fsgan}) and the two sets of genuine frames ({\orF} and {\orG}), where the subscripts indicate the dataset they belong.

%% ------------------------------------------------------ %%
\subsection{Frame category separation} 
As it will be discussed in Section \ref{sec:res}, we perform detection tests separately on two categories of frames: the ones where the face is fully shown and the ones where the subject places the hand/object in front of the face, indicated as {\woocc} and {\wocc}, respectively. This is functional for our experimental analysis (see Section \ref{ssec:cc}), as the sought occlusion-based artifacts in the swapped data appear only in the latter category. 

This separation is performed within all data partitions in a semi-automated manner for both datasets. 
For {\fows}, we applied the BlazeFace detector offered by the Google mediapipe framework \cite{mediapipe-face-detector} to locate the face bounding box in all video frames, which was then increased by a factor of 1.3 (as suggested in \cite{bbox_large, ff++}), to extract the whole user face. 
% and used to spatially crop all successive frames, since the prescribed face location is stable over the whole guiding video duration. 
Then, 100 frames are sampled from the static segment (see Figure \ref{fig:fows_ex}(a)) for the {\woocc} category; for the {\wocc} category, we run the BlazeFace detector with strict tolerance on the frames of dynamic segment, and selected 100 frames among the ones where such detector failed due to occlusion. 

%Once all user faces were detected,  \cite{mediapipe-face-detector} was applied again to separate between occluded and non-occluded challenges. Then, once all user frames were correctly organized, in order to create the occluded and non-occluded versions of the dataset by manually extracting the most informative frames. These are the central frames of the videos, where the user's face is completely occluded. 100 central frames were extracted for each video in the dataset. The same procedure was repeated for the non-occluded portion of the dataset, extracting 100 frames for each video, where the face was clearly visible. 
%In total, our dataset is composed by 4200 original frames (7 users recording 6 challenges each, we extracted 100 frames for each challenge), and 12600 swapped frames (7 users, each recording 6 challenges, we extracted 100 frames per challenge, and repeated the procedure for the 3 different face-swapping methods), for a total of 16.800 frames.
The latter procedure has also been used on the {\gotcha} dataset, which does not have separated static and dynamic segments. Therefore, the separation based on the BlazeFace detector has been applied on the whole temporal sequence.
For both datasets, a manual revision of the frames identified for each category was performed to refine the separation.

% \Note{rilasciare file con selezione video e separazione frames?}

% What we've done: 
% \begin{itemize}
%     \item computed total number of real and tot num of fakes 
%     \item took the smallest one as the number of images to extract for trn and tst set for the original subset
%     \item procedure:  tot\_imgs / 2 (we have 2 challenges) = num imgs per challenge
%     \item get the number of users for each split: num imgs per challenge / num users = num imgs per user to extract
%     \item divide this number by 2 since we are organizing each challenge in occ and no\_occ
%     \item get the number of expected imgs to extract
%     \item FineTuning part: noticed that the num of obj\_occ challenges were always less than the expected num of images to extract. We've decided to keep the number of obj occ fixed and increase the number of hand\_occ to extract by adding the difference between the extracted obj\_occ and the expected obj\_occ. This num was added to the num hand\_occ to extract and the process was repeated as before.
% \end{itemize}

\paragraph{Training/Testing split}

For {\fows}, data related to 5 users have been used for training, while the remaining 2 for testing. We consider the challenging setting where the two testing subjects are (i) the only woman in the dataset and (ii) the only man with a thick beard. %As discussed in \ref{dataset_collection}, our {\fows} dataset is composed by 7 users, the majority being male, and only one female. Among male users, only one with a thick beard stands out in terms of appearance. Consequently, the male and female subjects with complex appearance define the test set, while the remaining male users compose the training set. 
The 100 frames sampled from the dynamic and static segment during category separation yield 600 frames per user in each category, multiplied by 4 (the original version plus the three manipulated counterparts). Thus, for both {\wocc} and {\woocc} data, the training and testing splits are composed by 12000 (3000 original/ 9000 manipulated) and 4800 (1200 original/ 3600 manipulated) frames. 

% Table \ref{fig:fows_trntst} reports the Training/Testing split for our {\fows} dataset.

% The training/test split is obtained based on the user appearance. With the test split we are testing the model on the only female subject and on a male subject with a thick beard. We considered these two subjects to be more complex to recognize compared to the ones presented in the training split, composed of all remaining male users. 

% In order to define the training/test split in {\gotcha}, we 

In {\gotcha}, 30 users are selected for training and the remaining 17 for testing, making sure that no subject appears in both sets either as host or target face. %To reduce the number of {\fsgan} and {\dfl} swaps for each user, we assigned three random swap identities to each user and selected only the relative swaps.
% For each user, we selected three random identities used for both {\dfl} and {\fsgan} swaps.
%Once the training/test splits were defined, we have noticed that the original images were on average more than the fake images on all the dataset.
% In general, {\gotcha} provides more original than swapped samples, probably due to the fact that the fake videos were saved at a lower fps after processing. To compensate for this, we sample original frames for each user and each challenge, so as to have a comparable number of frames. The resulting split is reported in Table \ref{fig:gotcha_trntst}.
In general, {\gotcha} provides more original samples than swapped ones, probably due to the fact that fake videos were saved at lower fps after processing. To compensate for this, we sample original frames for each user and each challenge, to obtain a comparable number of frames. The resulting split is reported in Table \ref{fig:gotcha_trntst}.

\begin{table}[t!]
\caption{Numerosity of {\gotcha} training/testing splits.}
% \begin{minipage}{0.48\linewidth}
\scriptsize
\setlength{\tabcolsep}{1pt}
\centering
% \begin{tabular}{lccccc}
%   \cline{2-6}
%   \multicolumn{1}{l|}{} & 
%   \multicolumn{1}{c|}{\begin{tabular}[c]{@{}c@{}}{\orF}\\ trn | tst\end{tabular}} &
%   \multicolumn{1}{c|}{\begin{tabular}[c]{@{}c@{}}{\fd}\\ trn | tst\end{tabular}} &
%   \multicolumn{1}{c|}{\begin{tabular}[c]{@{}c@{}}{\gh}\\ trn | tst\end{tabular}} &
%   \multicolumn{1}{c|}{\begin{tabular}[c]{@{}c@{}}{\sims}\\ trn | tst\end{tabular}} &
%   \multicolumn{1}{c|}{\begin{tabular}[c]{@{}c@{}}Total \\ trn | tst\end{tabular}} \\ \hline
% \multicolumn{1}{|c|}{{\wocc}} &
%   \multicolumn{1}{c|}{3000 | 1200} &
%   \multicolumn{1}{c|}{3000 | 1200} &
%   \multicolumn{1}{c|}{3000 | 1200} &
%   \multicolumn{1}{c|}{3000 | 1200} &
%   \multicolumn{1}{c|}{12000 | 4800} \\ \hline
% \multicolumn{1}{|c|}{{\woocc}} &
%   \multicolumn{1}{c|}{3000 | 1200} &
%   \multicolumn{1}{c|}{3000 | 1200} &
%   \multicolumn{1}{c|}{3000 | 1200} &
%   \multicolumn{1}{c|}{3000 | 1200} &
%   \multicolumn{1}{c|}{12000 | 4800} \\ \hline
%   %% just to have tables aligned
%  % &
%  %  \multicolumn{1}{l}{} &
%  %  \multicolumn{1}{l}{} &
%  %  \multicolumn{1}{l}{} &
%  %  \multicolumn{1}{l}{} &
%  %  \multicolumn{1}{l}{} \\
%  %  &
%  %  \multicolumn{1}{l}{} &
%  %  \multicolumn{1}{l}{} &
%  %  \multicolumn{1}{l}{} &
%  %  \multicolumn{1}{l}{} &
%  %  \multicolumn{1}{l}{} \\
% \end{tabular}
% % \space
% % \\
% \caption{{\fows} dataset Training/Testing split.}
% \label{fig:fows_trntst}
% \end{minipage}%

% \hfill
% \begin{minipage}{0.48\linewidth}
% \scriptsize
% \setlength{\tabcolsep}{1pt}
% \centering
\begin{tabular}{l|c|cc|c}
%\cline{2-5}
 &
  \begin{tabular}[c]{@{}c@{}}{\orG}\\ trn | tst\end{tabular} &
  \multicolumn{1}{c|}{\begin{tabular}[c]{@{}c@{}}{\dfl}\\ trn | tst\end{tabular}} &
  \begin{tabular}[c]{@{}c@{}}{\fsgan}\\ trn | tst\end{tabular} &
  \begin{tabular}[c]{@{}c@{}}Total \\ trn | tst\end{tabular} \\ \hline
\multicolumn{1}{c|}{{\wocc}} &
  {28395| 19463} &
  \multicolumn{1}{c|}{13716| 9899} &
  {14057| 10019} &
  {56168| 39381} \\ \hline
\multicolumn{1}{c|}{{\woocc}} &
  15188 | 13200 &
  \multicolumn{1}{c|}{7922 | 6338} &
  {7906 | 6407} &
  {31016 | 25945} \\ %\hline
%\multicolumn{1}{|l|}{\# frames per algo} &
%  \multicolumn{1}{l|}{43583 | 32663} &
%  \multicolumn{1}{l|}{21638 | 16237} &
%  \multicolumn{1}{l|}{21963 | 16426} &
%  \multicolumn{1}{l|}{} \\ \hline
%\multicolumn{1}{|l|}{\# frames per class} &
%  \multicolumn{1}{l|}{43583 | 32663} &
%  \multicolumn{2}{c|}{43601 | 32663} &
%  \multicolumn{1}{l|}{} \\ \hline 
\end{tabular}

\label{fig:gotcha_trntst}
% \end{minipage}%
\end{table}

\subsection{Detection models}

% We consider a battery of five CNN architectures, most of them previously used on face video manipulation detection \cite{ff++}. While we are aware that more sophisticated approaches exist, the analysis in \cite{DFB} show that such general-purpose architectures (especially the ones involving depthwise separable convolution components) provide comparable performance to handcrafted approaches, in front of a reduced complexity. 
We consider a battery of five CNN architectures, most of them previously used on face video manipulation detection \cite{ff++}. While we are aware that more sophisticated approaches exist, the analysis in \cite{DFB} shows that such general-purpose architectures provide performance comparable to handcrafted approaches, in front of a reduced complexity. 
We therefore consider three baseline models pretrained on ImageNet: MobileNetv2 ({\mnet})%\cite{mnetv2}
, EfficientNetB4 ({\enet})%\cite{effnetb4}
, and XceptionNet ({\xnet}) \cite{xception}. Moreover, we include two CNNs already specialized in deepfake detection:  
\begin{itemize}
    \item {\enetPRE}: proposed in \cite{MilanPaper}, based on the EfficientNetB4 architecture and trained on the DFDC dataset;
    \item {\xnetPRE}: used in \cite{DFB}, based on the Xception architecture and trained on FaceForensics++.
\end{itemize}

Both models are selected due to their performance in the respective papers. We use the original versions released by the authors \cite{MilanPaper} \cite{DFB} and replicate their data preprocessing to ensure fairness in the evaluation.
%The selected deepfake detectors, instead, are pretrained on state of the art deepfake detection datasets. The EffcientNetB4 ({\enetPRE}) in \cite{MilanPaper} is pretrained on DFDC, while the Xception ({\xnetPRE}) model in \cite{DFB} is pretrained on FF++.
% \begin{itemize}
%     \item {\enetPRE}: EfficientNetB4 as in \cite{MilanPaper} (pretrained on DFDC)
%     \item {\xnetPRE}: Xception as in \cite{DFB} (pretrained on FF++)
% \end{itemize}
%To develop the training and testing dataset, the selected deepfake models apply preprocessing strategies that are similar to what we have described in section \ref{dataset_collection}. 
%In \cite{MilanPaper}, the authors apply the BlazeFace detector (the same we used to detect faces with google's mediapipe \cite{mediapipe-face-detector}), to extract each face from the frames. The detected faces are resized to (224, 224) pixels. 
%Instead, in \cite{DFB}, the authors use dlib (cite dlib) to detect faces in videos. All detected faces are then aligned and cropped by enlarging the detected face by 1.3 times to include more of the face region (similarly to what we have done in developing our dataset). All detected faces are then resized to (256, 256) pixels. 

\subsection{Testing protocol}

We adopt a binary classification framework: the frames in {\orF} and {\orG} are associated to the label 0, the manipulated frames in the other partitions are associated to the label 1. Classification is performed according to the default 0.5 threshold on the CNN output score.
% removed pythorc implementation of focal loss
All CNN models are retrained on both {\fows} and {\gotcha} datasets, separately.
For the baselines, we use the {\it focal loss} function \cite{focal-loss}, a modified version of the Cross Entropy loss that adaptively focuses the model learning on less represented samples, to counter the class imbalance between original and manipulated data in the {\fows} dataset.
% [original version of the sentence] We applied the Adam optimizer, as in \cite{DFB} \cite{MilanPaper}
A simple early-stopping procedure (with patience 3) was adopted to control overfitting during training. Models are trained for 10 epochs on {\fows}, using the Adam optimizer as in \cite{DFB} \cite{MilanPaper}, and for 15 epochs on {\gotcha}, applying the AdamW optimizer \cite{loshchilov2018decoupled}. 
After some preliminary tests (not reported for the sake of space), we identified the best performing PyTorch data augmentation configuration as follows: random resize crop at (224,224), random horizontal flip, rotation in the range [-5,5] degrees, and color jitter with default values.
%In training, we applied the following PyTorch data augmentations to concentrate the model on the central portion of the face, where the most information is located. We performed some preliminary data augmentation experiments (called {\it base\_aug}) on the \mnet model. The data augmentations that provide better results in terms of deepfakes detection are the following: 
% and also should help the model's robustness in face detection since they extract random crops of the face. 
% During training we wanted to apply data augmentation to our data. In particular, we wanted the model to concentrate on the center of the face, where the most information is located. Therefore, we applied the following PyTorch data augmentations:
%\begin{itemize}
    % \item Resize the image to (256, 256)
    %% from preliminary experiments on Mobilenet these augmentations provided better resutls also should help robusteness in face detection since it extracts random crop of the face. 
%    \item Random Resized Crop, which extracts a random crop of the image and resizes it to the desired size.
%    \item Random Horizontal Flip 
%    \item Random Rotation between [-5, 5] degrees
%    \item ColorJitter, randomly adjust the brightness, contrast, saturation, and hue of the image (we applied the default values)
%\end{itemize}
For the {\enetPRE} and {\xnetPRE} CNNs, we used the same loss and augmentation configurations as in the original code. 
%Both {\enetPRE} \cite{MilanPaper} and {\xnetPRE} \cite{DFB} were trained on {\fows} for 10 epochs and using the Focal Loss. However, when trained on {\gotcha}, only the {\enetPRE} model was trained following the same procedure describe for the baseline, but applying the BCE with logit loss defined in the paper.
%In the case of {\xnetPRE} \cite{DFB}, when trained on {\gotcha}, we were unable to load the whole data set into training. Thus, to obtain a manageable size of the data set, we used the training strategy defined by the authors. So, the model was trained for 10 epochs, using the CE loss as loss function, and from each video only 32 frames were extracted. As investigated in \cite{DFB} and \cite{MilanPaper}, extracting 32 frames from each video helps avoid overfitting when training the model. 
% To replicate the same metric used by the baseline models during validation, in the {\xnetPRE} we set the {metric\_scoring} to {acc} instead of {auc}. 

% {\it Describe training details: retrain all layers, use of focal loss, which augmentation is applied, descent algorithm, epoch, stopping criteria, etc}

\subsection{Metrics}
%% removed sci-kit learn BA definition
We applied different metrics for a comprehensive evaluation. Based on the model decisions, we report the {\it Balanced Accuracy} ({\ba}), defined as the average between sensitivity and specificity, which is typically used in case of imbalanced datasets and provides more representative values with respect to the overall accuracy; we also report the accuracy {\acc{$p$}} separately for each partition $p$ defined in Section \ref{ssec:data}.
Moreover, we compute the {\it Area Under Curve ({\auc})}, obtained by binarizing the network output with different thresholds, and the {\it Equal Error Rate ({\eer})}.

%  \cite{ba-scikit}

% \todo{add brief description of the metrics}

% We do not report the overall accuracy of the model since it was usually imbalanced due to how the dataset were defined

% {\it Metrics: BA, ACC$_{partition}$, AUC, EER}
% \\
% {\todo{BA e ACC devono essere riportate con la stessa scala: o percentuale o da 0 a 1; AUC e EER sempre [0,1]}
% \todo{Acc overall non riportata perché troppo sbilanciata, riportata BA per avere risultati più bilanciati}

\section{Results}
\label{sec:res}

%\todo{Check performance of Xception-DFB in the aligned setting and generalization; distribution of scores in the cross-dataset case}

% \todo{add model hist in cases of high auc and low acc}
% \Note{mobnet/xception-gotcha-occ vs fows-occ \\ mobnet/xception-gotcha-no-occ vs fows-no-occ}

We first analyze model performance in a fully aligned setting, i.e., where train and test data belong to the same dataset and frame category. In Table \ref{tab:aligned}, results for {\fows} and {\gotcha} are arranged column-wise, while the top (bottom) row refers to the {\wocc} (\woocc) category.

It can be observed that most models yield excellent results in all cases. For {\gotcha} in particular, the distinction is essentially perfect for both the {\wocc} and {\woocc} categories.
We observe that {\enetPRE} and {\xnetPRE} do not obtain superior results with respect to their baseline versions, {\enet} and {\xnet}, on the {\woocc} data. In other words, it emerges that being previously trained on swapped data from other datasets (DFDC and FF++) does not bring an advantage in dealing with swapped data coming from different sources ({\fows} and {\gotcha}). 
This holds in particular for {\xnetPRE}, for which retraining on {\fows} (where less retraining data are available with respect to {\gotcha}) does not lead to accurate results. This shows a general difficulty of detectors in effectively capturing discriminative cues for data subject to equivalent manipulations (face swapping with no occlusion) but coming from different data sources, as also observed in \cite{DFB}.

% \acc{\orG}

% \begin{table*}[t!]
%     \centering
%     \includegraphics[width=0.8\linewidth]{imgs/res-aligned.png}
%     \caption{Model performance in the fully aligned setting. \todo{Riportare i risultati in tabella latex con la notazione definita nelle sezioni 2 e 3.}}
%     \label{tab:aligned}
% \end{table*}
%% --------------------------------------------------------- %%
%% ALIGNED-DATASET TABLE %%
\begin{table*}[t!]
\caption{Model performance in the fully aligned setting.}
% \begin{subtable}[h]{0.35\linewidth}
\begin{minipage}{0.48\linewidth}
\scriptsize
\setlength{\tabcolsep}{1pt}
\centering
\begin{tabular}{@{}lccccccc@{}}
% \centering
\toprule
% \rowcolor[HTML]{FFFFFF} 
\multicolumn{8}{c}{Train: {\fows}, {\wocc}. Test: {\fows}, {\wocc}.}  \\ \midrule
% \rowcolor[HTML]{FFFFFF} 
\multicolumn{1}{l}{\textbf{Model}} &
\multicolumn{1}{c}{\textbf{{\ba}}} &
\multicolumn{1}{c}{\textbf{\acc{\orF}}} &
\multicolumn{1}{c}{\textbf{\acc{\sims}}} &
\multicolumn{1}{c}{\textbf{\acc{\gh}}} &
\multicolumn{1}{c}{\textbf{\acc{\fd}}} &
\multicolumn{1}{c}{\textbf{{\auc}}} &
\multicolumn{1}{c}{\textbf{{\eer}}} \\
{\mnet} & 99.56 & 99.17 & 99.92 & 100 & 99.92 & 1.0000 & 0.0033 \\
% {\mnet}\_baseAug   & 84.59 & 69.17 & 100 & 100 & 100 & 1.00 & 0.0142 \\
{\enet}          & 97.88 &95.75 & 100 &	100 & 100 & 0.9999 & 0.0050 \\
% {\enet}\_baseAug & 83.92 & 67.83 & 100 & 100 & 100 & 1.00 & 0.0158 \\
{\xnet}          & 99.10 & 98.25 & 99.83 & 100 & 100 & 0.9999 & 0.0033 \\
% {\xnet}\_baseAug & 93.25 & 86.50 & 100 & 100 & 100 & 1.00 & 0.0025 \\
{\enetPRE}          & 99.05 & 98.92 & 99.92 & 100 &	97.58 &	0.9995 & 0.0100 \\
{\xnetPRE}       & 71.03 & 46.83 & 91.42 & 100 & 94.25 & 0.8472 & 0.2700 \\ 
\bottomrule
% {Median {\ba}} & 99.05 \\ \bottomrule
\end{tabular}

\end{minipage}%
\begin{minipage}{0.48\linewidth}%
\scriptsize
\setlength{\tabcolsep}{1pt}
\centering
\begin{tabular}{@{}lccccccc@{}}
\toprule
\multicolumn{7}{c}{Train: {\gotcha}, {\wocc}. Test: {\gotcha}, {\wocc}.} & \multicolumn{1}{l}{} \\ \midrule
\textbf{Model} &
  \multicolumn{1}{c}{\textbf{{\ba}}} &
  \multicolumn{1}{c}{\textbf{\acc{\orG}}} &
  \multicolumn{1}{c}{\textbf{\acc{\dfl}}} &
  \multicolumn{1}{c}{\textbf{\acc{\fsgan}}} &
  \multicolumn{1}{c}{\textbf{{\auc}}} &
  \multicolumn{1}{c}{\textbf{{\eer}}} \\
{\mnet}    & 99.96  & 99.91  & 100 & 100 & 1.0000 & 0.0001 \\
{\enet} & 99.99 & 100 & 100 & 99.95 & 1.0000 & 0.0000 \\
{\xnet}    & 99.98 & 99.96 & 100 & 100 & 1.0000 & 0.0001 \\
{\enetPRE}     & 99.97 & 99.96 & 100 & 99.96 & 1.0000 &	0.0003 \\
{\xnetPRE}        & 99.76 &	99.51 &	100 & 100 &	1.0000 & 0.0029 \\
\bottomrule
% {Median {\ba}} & {99.97} \\\bottomrule
\end{tabular}
\end{minipage}%

\begin{minipage}{0.48\linewidth}%
\scriptsize
\setlength{\tabcolsep}{1pt}
    \centering
    \begin{tabular}{@{}lccccccc@{}}
        % \toprule
        \\
        \toprule
        \multicolumn{8}{c}{ Train: {\fows}, {\woocc}. Test: {\fows} {\woocc}.}  \\ \midrule
        \multicolumn{1}{l}{\textbf{Model}} &
          \multicolumn{1}{c}{\textbf{{\ba}}} &
          \multicolumn{1}{c}{\textbf{\acc{\orF}}} &
          \multicolumn{1}{c}{\textbf{\acc{\sims}}} &
          \multicolumn{1}{c}{\textbf{\acc{\gh}}} &
          \multicolumn{1}{c}{\textbf{\acc{\fd}}} &
          \multicolumn{1}{c}{\textbf{{\auc}}} &
          \multicolumn{1}{c}{\textbf{{\eer}}} \\
        {\mnet} & 100 & 100 & 100 & 100 & 100 & 1.0000 & 0.0000 \\
        {\enet} & 100 & 100 & 100 & 100 & 100 & 1.0000 & 0.0000 \\
        {\xnet} & 99.92  & 100 & 100 & 100 & 100 & 1.0000 & 0.0000 \\
        {\enetPRE} & 95.72  & 99.00  & 99.83  & 99.83  & 77.67  & 0.9965 & 0.0325 \\
        {\xnetPRE} & 66.51  & 74.33  & 51.17  & 53.08  & 71.83  & 0.7368 & 0.3392 \\ 
        \bottomrule
        % {Median {\ba}} & {99.92} \\
        % \bottomrule
    \end{tabular}
\end{minipage}%
\begin{minipage}{0.48\linewidth}%
\scriptsize
\setlength{\tabcolsep}{1pt}
\centering
\begin{tabular}{@{}lccccccc@{}}
\\
\toprule
\multicolumn{7}{c}{Train: {\gotcha}, {\woocc}. Test: {\gotcha}, {\woocc}.} & \multicolumn{1}{l}{} \\ \midrule
\textbf{Model} &
  \multicolumn{1}{c}{\textbf{{\ba}}} &
  \multicolumn{1}{c}{\textbf{\acc{\orG}}} &
  \multicolumn{1}{c}{\textbf{\acc{\dfl}}} &
  \multicolumn{1}{c}{\textbf{\acc{\fsgan}}} &
  \multicolumn{1}{c}{\textbf{{\auc}}} &
  \multicolumn{1}{c}{\textbf{{\eer}}} \\
{\mnet}    & 100 & 99.76   & 100 & 100  & 1.0000 & 0.0013 \\
{\enet} & 100 & 100 & 100 & 100 & 1.0000 & 0.0000 \\
{\xnet}    & 100 & 99.21 & 100 & 100 & 0.9998 & 0.0020 \\
{\enetPRE}      & 100 & 99.99 & 100 & 100 & 1.0000 & 0.0000 \\
{\xnetPRE}        & 100 & 100 & 100 & 100 & 1.0000 & 0.0000 \\ 
\bottomrule 
% {Median {\ba}} & {100} \\
% \bottomrule
\end{tabular}
\end{minipage}%

\label{tab:aligned}
\end{table*}
%% --------------------------------------------------------- %%

% \todo{esplicitare train/test nel'header; uniformare le cifre significative nelle diverse tabelle (due dopo la virgola per le acc e 4 per auc e err}; quando è 100 togliere i decimali}

\subsection{Cross-dataset tests}
% In order to assess their generalization capabilities, detection models are now tested in a cross-dataset fashion, thus performing training on one dataset and testing on the other one.
In order to assess their generalization capabilities, detection models are tested in a cross-dataset fashion, thus training on one dataset and testing on the other. Table \ref{tab:cd} reports the results for both categories, where right and left blocks refer to the training dataset.
Here, the accuracy metrics show a significant drop in performance for all settings. The {\ba} almost halves for most models, and accuracy values on the different dataset partition reveals poor capabilities of correctly classifying video frames.
We observe that the prominent type of misclassification error changes together with the testing dataset: models trained on {\gotcha} tend to miss manipulated frames on {\fows}, while models trained on {\fows} are prone to classifying genuine {\gotcha} frames as manipulated. While not reporting here the full results for the sake of space, we stress that the aligned and cross-dataset setting tests have also been replicated in a transfer learning mode (by training the last layer only instead of the whole network) for an extended evaluation. 
The results obtained are consistent: the median {\ba} over the models in the aligned setting is equal to 90\%, while it drops to 65\% in the cross-dataset setting.

\begin{figure}[ht!]
    \centering

    \begin{subfigure}{\linewidth}
    \centering
        \includegraphics[width = 0.49\textwidth]{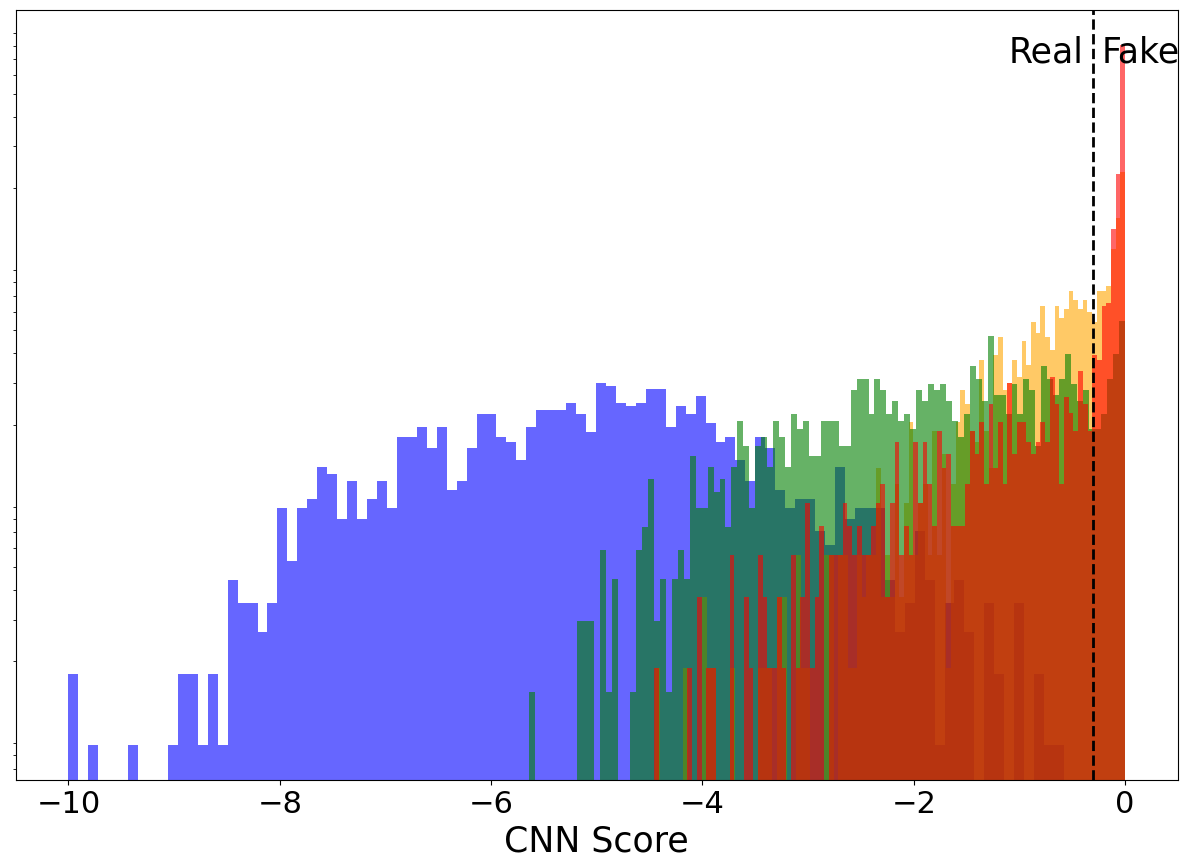}
        \includegraphics[width = 0.49\textwidth]{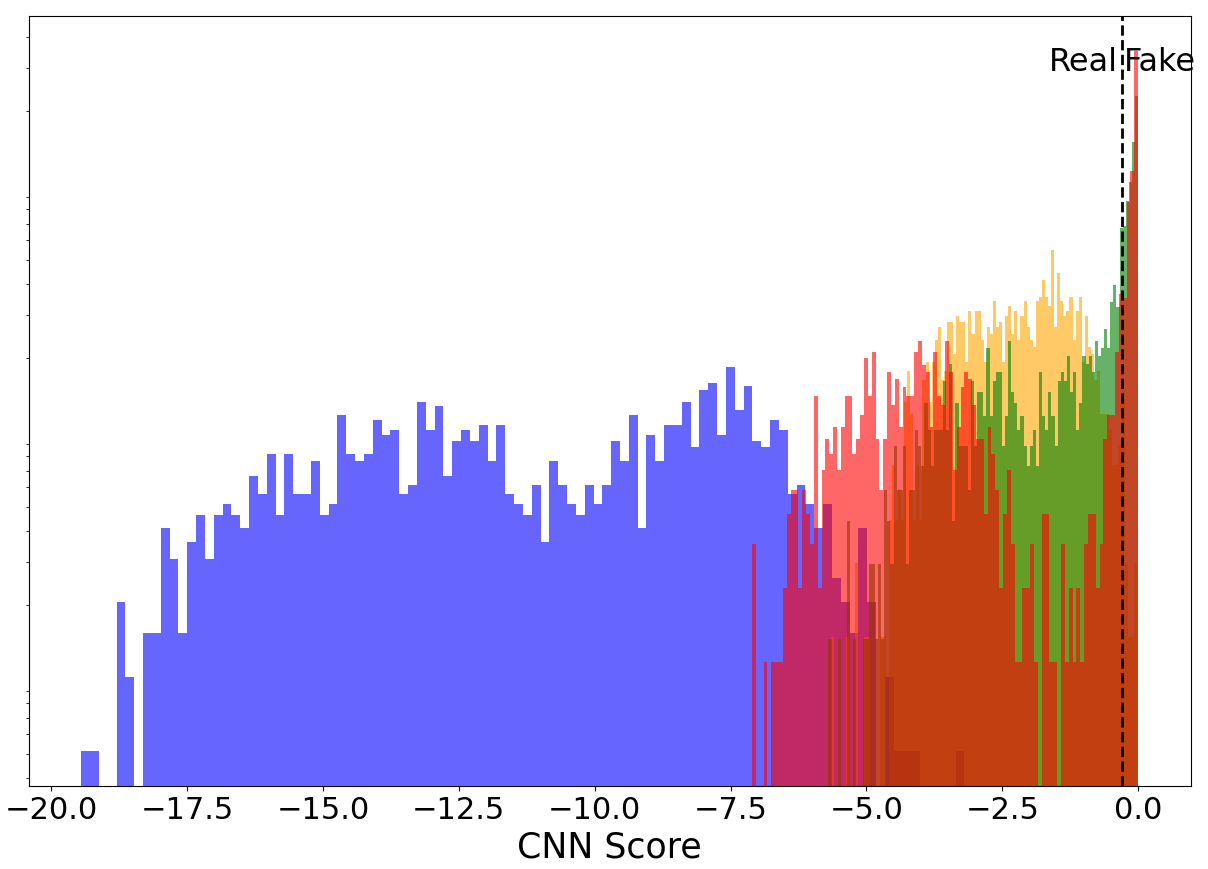}
        
    \end{subfigure}

    % \begin{minipage}{\linewidth}
    % \centering
    %     \includegraphics[width = 0.85\linewidth]{imgs/prob_hist/_BIG_MnetV2_GOTCHA_wocc_FT_vs_FOWS_wocc_hist.png}
    % \end{minipate}

    \includegraphics[width = 0.4\textwidth]{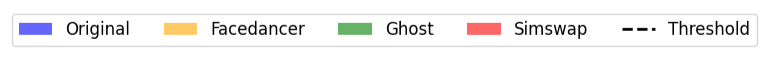}

    % \begin{subfigure}{\linewidth}
    % \centering
    %     \includegraphics[width = 0.65\linewidth]{imgs/prob_hist/_BIG_Xception_gotcha_woocc_FT_vs_thesis_woocc_hist_.png}
    % \end{subfigure}
    % \includegraphics[width=0.24\textwidth]{imgs/prob_hist/_MnetV2_gotcha_wocc_FT_vs_thesis_wocc_prob_hist_0.5_thresh.png}
    % \includegraphics[width=0.24\textwidth]{imgs/prob_hist/_Xception_gotcha_woocc_FT_vs_thesis_woocc_hist__prob_hist_0.5_thresh.png}

    \caption{Histogram of the decision score for {\mnet} (left) and {\xnet} (right) trained {\gotcha} and tested on {\fows}. 
    }
    \label{fig:prob-hist}
\end{figure}

\begin{figure}[ht!]
\centering
\begin{subfigure}{\linewidth}
\centering
\includegraphics[width=0.9\linewidth]{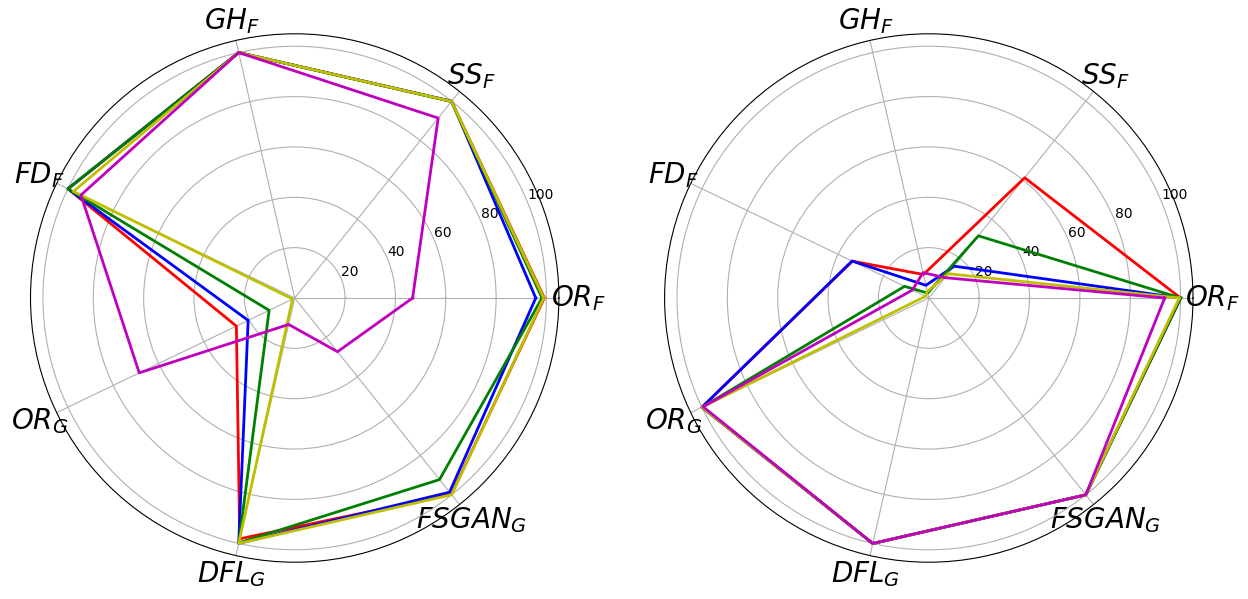}
% \end{subfigure}

    % \includegraphics[width=0.5\linewidth]{imgs/sp/spider_plot_combined_2.png}

%     % (a)
\end{subfigure}

\begin{subfigure}{\linewidth}
    \centering
    \includegraphics[width=0.9\linewidth]{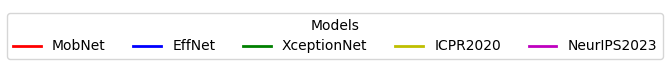}
\end{subfigure}%

\caption{ Visualization of the accuracy on all partitions of the models trained on {\fows} (left), and the ones trained on {\gotcha} (right). 
}
\label{radar-charts}
\end{figure}

An interesting observation is that the strong performance decrease in terms of accuracy is not always reflected in the {\auc} and {\eer} metrics: when training on {\gotcha} and training on {\fows}, models like {\mnet} and {\xnet} retain rather good values of {\auc} and {\eer} for both categories. This indicates that the model distinguishes genuine from manipulated samples in terms of prediction scores, but does not classify them correctly with the default threshold on 0.5, which was instead effective on the training dataset.
Figure \ref{fig:prob-hist} reports the score distributions (in logarithmic scale) in those cases, showing that there is indeed a separation but approaching zero. 
While denoting a certain generalization capabilities, this shift on the score distribution represents a relevant issue in practical situations where unseen data (potentially coming from different data sources) would be tested through a pretrained model, as applying the prescribed threshold would lead to strong inaccuracies. The charts in Figure \ref{radar-charts} visualize this scenario, showing that no model behaves reliably on all partitions.

%% ---------------------------------------------------------- %%
%% Prob histograms %%
%% xception_gotcha_woocc_fows_woocc
%% mnet_gotcha_wocc_fows_wocc
%% --------------------------------------------------------- %%
%% SPIDER PLOTS %%

%% --------------------------------------------------------- %%
%% CROSS-DATASET TABLE %%
\begin{table*}[t!]
\caption{Model performance in the cross-dataset setting.}
%% fows no_occ vs gotcha no_occ
\begin{minipage}{0.48\linewidth}%
\scriptsize
\setlength{\tabcolsep}{1pt}
\centering
    \begin{tabular}{@{}lcccccc@{}}
\toprule
\multicolumn{7}{c}{Train: {\fows}, {\wocc}. Test: {\gotcha}, {\wocc}.} \\ \midrule
\multicolumn{1}{l}{\textbf{Model}} &
  \multicolumn{1}{c}{\textbf{{\ba}}} &
  \multicolumn{1}{c}{\textbf{\acc{\orF}}} &
  \multicolumn{1}{c}{\textbf{\acc{\dfl}}} &
  \multicolumn{1}{c}{\textbf{\acc{\fsgan}}} &
  \multicolumn{1}{c}{\textbf{{\auc}}} &
  \multicolumn{1}{c}{\textbf{{\eer}}} \\
{\mnet} & 62.34 & 25.72 & 98.12  & 99.77 & 0.7120 & 0.3135 \\
{\enet}  & 59.92 & 20.52 & 99.98  & 98.66 & 0.7833 & 0.2560 \\
{\xnet} & 53.64 & 11.27 & 99.80  & 92.26 & 0.6181 & 0.3968 \\
{\enetPRE} & 50.46 & 0.94  & 100 & 99.96 & 0.8555 & 0.2153 \\
{\xnetPRE}& 43.79 & 68.55 & 10.78  & 27.29   & 0.3901 & 0.5703 \\ 
\bottomrule
% {Median \ba} & {53.64} \\
% \bottomrule
\end{tabular}
\end{minipage}
\begin{minipage}{0.48\linewidth}
\scriptsize
\setlength{\tabcolsep}{1pt}
\centering
% gotcha_occ s fows_occ
\begin{tabular}{@{}lccccccc@{}}
\toprule
\multicolumn{7}{c}{ Train: {\gotcha}, {\wocc}. Test: {\fows}, {\wocc}.} \\ \midrule
  \multicolumn{1}{l}{\textbf{Model}} &
  \textbf{{\ba}} &
  \textbf{\acc{\orF}} &
  \textbf{ {\acc{\sims}}} &
  \textbf{\acc{\gh}} &
  \textbf{\acc{\fd}} &
  \textbf{{\auc}} &
  \textbf{\eer} \\
% {\gotcha} {\wocc} & 
{\mnet} & 67.39 & 100 & 61.08 & 9.58  & 33.67 & 0.9725 & 0.0942 \\
 {\enet} & 59.20 & 100   & 16.17 & 5.17 & 33.83 & 0.9243 & 0.1675 \\
{\xnet} & 57.38 & 100 & 31.58 & 2.00  & 10.67 & 0.9405 & 0.1442 \\
{\enetPRE} & 52.50 & 99.50  & 12.33 & 2.67  & 1.50 & 0.7630 & 0.3075 \\
{\xnetPRE} & 51.49 & 93.67  & 10.33 & 10.33 & 7.25    & 0.4736 & 0.5375 \\ \bottomrule 
% {Median {\ba}} & {57.38} \\
% \bottomrule
\end{tabular}
\end{minipage}%
\\
%% fows_no_occ vs gotcha_no_occ
\begin{minipage}{0.48\linewidth}
\scriptsize
\setlength{\tabcolsep}{1pt}
\centering
    \begin{tabular}{@{}lcccccc@{}}
% \toprule
\\
\toprule
\multicolumn{7}{c}{Train: {\fows}, {\woocc}. Test: {\gotcha}, {\woocc}.} \\ \midrule
\multicolumn{1}{l}{\textbf{Model}} &
  \multicolumn{1}{c}{\textbf{{\ba}}} &
  \multicolumn{1}{c}{\textbf{\acc{\orF}}} &
  \multicolumn{1}{c}{\textbf{\acc{\dfl}}} &
  \multicolumn{1}{c}{\textbf{\acc{\fsgan}}} &
  \multicolumn{1}{c}{\textbf{{\auc}}} &
  \multicolumn{1}{c}{\textbf{{\eer}}} \\
{\mnet} & 53.48 & 7.08  & 99.81  & 99.95  & 0.6552 & 0.3784 \\
{\enet} & 51.98 & 3.95  & 100 & 100 & 0.8461 & 0.2113 \\
{\xnet} & 54.14 & 8.704 & 99.15  & 100 & 0.6459 & 0.3887 \\
{\enetPRE} & 51.23 & 4.10  & 99.15  & 97.58  & 0.6708 & 0.3448 \\
{\xnetPRE} & 31.41 & 25.57 & 15.35  & 59.11  & 0.2078 & 0.7363 \\ 
\bottomrule
% {Median \ba} & {54.98} \\
% \bottomrule
\end{tabular}
\end{minipage}%
%% gotcha_occ vs fows_occ
\begin{minipage}{0.48\linewidth}%
\scriptsize
\setlength{\tabcolsep}{1pt}
\centering
    \begin{tabular}{@{}lccccccc@{}}
\\
\toprule
\multicolumn{8}{c}{Train: {\gotcha}, {\woocc}. Test: {\fows}, {\woocc}.} \\ \midrule
\multicolumn{1}{l}{\textbf{Model}} &
  \multicolumn{1}{c}{\textbf{{\ba}}} &
  \multicolumn{1}{c}{\textbf{\acc{\orG}}} &
  \multicolumn{1}{c}{\textbf{\acc{\sims}}} &
  \multicolumn{1}{c}{\textbf{\acc{\gh}}} &
  \multicolumn{1}{c}{\textbf{\acc{\fd}}} &
  \multicolumn{1}{c}{\textbf{{\auc}}} &
  \multicolumn{1}{c}{\textbf{{\eer}}} \\
{\mnet}& 55.00 & 99.83 & 25.50 & 4.67 & 0.25 & 0.9462 & 0.1250 \\
{\enet} & 78.00 & 100 & 79.08  & 56.67  & 32.00  & 0.9609 & 0.1258 \\
{\xnet}& 64.00 & 100 & 42.33  & 37.75 & 1.25  & 0.9982 & 0.0225 \\
{\enetPRE} & 50.00 & 100 & 0.00  & 0.00  & 0.00  & 0.5969 & 0.4350 \\
{\xnetPRE}& 47.99 & 77.33  & 5.58  & 27.33 & 23.00 & 0.4557 & 0.5183 \\ 
\bottomrule
% {Median {\ba}} & {55.00} \\
% \bottomrule
\end{tabular}
\end{minipage}%

    \label{tab:cd}
\end{table*}
%% --------------------------------------------------------- %%

% {\it 

% %Visualization of Gradcam results
% %\\ \TBD{Missing EffNetB4 and Xception baselines GradCAMs}

% }

\subsection{Cross-category tests}
\label{ssec:cc}
As widely recognized in the deep learning domain, a crucial issue in the training and deployment of deep networks is the interpretability of the discriminative cues they learn. In order to have further insights on our detection scenario, we have performed experiments where models trained on data belonging to a certain category type are tested on the other category of the same dataset. 
In fact, the {\wocc} frames contain visual artifacts caused by face occlusion while those are not present in the {\woocc} frames. Table \ref{tab:cs} reports the results in two insightful settings: the first one explores the gap in terms of EER on {\wocc} when training on {\wocc} or {\woocc}; the second one assesses detection performance when a model trained on {\wocc} is tested on {\woocc} instead of {\wocc} data.

In both cases, we see a rather limited decrease in terms of EER when moving from an aligned to a non-aligned setting. In particular, it is shown that the {\wocc} data are well separated by both {\wocc}- and {\woocc}-trained models, which have not seen occlusion-based artifacts in training (top subtable). Also, a {\wocc}-trained model performs well also on {\woocc}, thus in absence of occlusion-based artifacts (bottom subtable).
This suggests that the decisions of {\wocc}-trained network are only marginally based on actual occlusion artifacts, while they likely learn shared discriminative cues among the two frame categories.

% For further insights, -> Lastly
Lastly, we investigate the discriminative features the network focuses on by applying \textbf{GradCAM++} \cite{pytorchcam}. For the sake of space, we report the results for the {\mnet} and {\enetPRE} models (yielding superior performance in Table \ref{tab:cd}) tested on sample swapped frames. Looking at the results, we can see that the {\mnet} activations vary over the test data, but they are often not concentrated on the occluded part where the visual artifacts are actually located. On the other hand, {\enetPRE} activations are consistently located in the face center, thus missing other visually relevant areas.

\begin{table}[h!]
\caption{{\eer} values for different combinations of training/testing frame categories.}
    % \centering
    % \includegraphics[width=0.45\textwidth]{imgs/res-cs.png}
    % \caption{{\eer} values for different combinations of training/testing segment types.}
    % \label{tab:cs}
\begin{minipage}{\linewidth}
\centering
\begin{tabular}{@{}lrr|rr@{}}
& \multicolumn{2}{c}{{\fows}}      & \multicolumn{2}{c}{{\gotcha}}   \\ \toprule
Test & \multicolumn{4}{c}{{\wocc}}  %& \multicolumn{2}{c}{{\wocc}}  
\\ \midrule
Train  & \multicolumn{1}{l}{{\wocc}}   & \multicolumn{1}{l|}{{\woocc}} & \multicolumn{1}{l}{{\wocc}}   & \multicolumn{1}{l}{{\woocc}} \\ \midrule
{\mnet}       & 0.00 & 0.04  & 0.00 & 0.00 \\
{\enet}     & 0.01 & 0.01  & 0.00 & 0.00 \\
{\xnet}     & 0.00 & 0.00  & 0.00 & 0.00 \\
{\enetPRE} & 0.01 & 0.20  & 0.00 & 0.00 \\
{\xnetPRE} & 0.27 & 0.39  & 0.00 & 0.00
\end{tabular}
\end{minipage}%

\vspace{0.4cm}

\begin{minipage}{\linewidth}
\centering
\begin{tabular}{@{}lrr|rr@{}}
& \multicolumn{2}{c}{{\fows}}      & \multicolumn{2}{c}{{\gotcha}}   \\ \toprule
Test  & \multicolumn{1}{l}{{\wocc}}   & \multicolumn{1}{l|}{{\woocc}} & \multicolumn{1}{l}{{\wocc}}   & \multicolumn{1}{l}{{\woocc}} \\ \midrule
Train & \multicolumn{4}{c}{{\wocc}}  %& \multicolumn{2}{c}{{\wocc}}  
\\ \midrule
{\mnet}       & 0.00 & 0.0000  & 0.00 & 0.00 \\
{\enet}     & 0.01 & 0.0000 & 0.00 & 0.00 \\
{\xnet}     & 0.00 & 0.0000  & 0.00 & 0.00 \\
{\enetPRE} & 0.01 & 0.0000  & 0.00 & 0.00 \\
{\xnetPRE} & 0.27 & 0.2067  & 0.00 & 0.00
\end{tabular}
\end{minipage}

\label{tab:cs}
\end{table}

% \todo{Gradcams in these settings}
% \Note{2 rows of 4 imgs of working/not working examples for occ case}
\begin{figure}[ht!]
% \centering
% \begin{minipage}{0.48\linewidth}
\begin{subfigure}{\linewidth}
\centering
    % \begin{subfigure}[b]{0.24\textwidth}
        
            \includegraphics[width=0.2\linewidth]{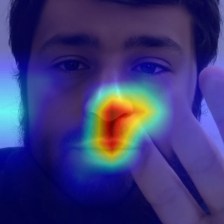}
            \includegraphics[width=0.2\linewidth]{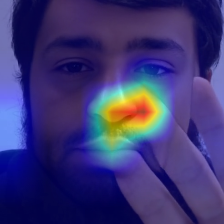}
            \includegraphics[width=0.2\linewidth]{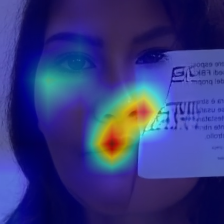}
            \includegraphics[width=0.2\linewidth]{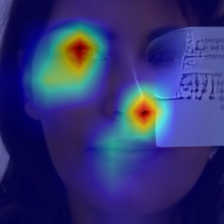}
\end{subfigure}
\begin{subfigure}{\linewidth}
\centering
            \includegraphics[width=0.2\linewidth]{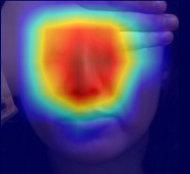}
        \includegraphics[width=0.2\linewidth]{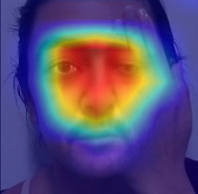}
        \includegraphics[width=0.2\linewidth]{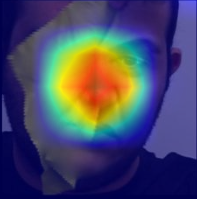}
        \includegraphics[width=0.2\linewidth]{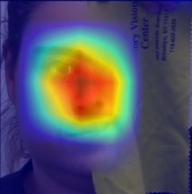}

\end{subfigure}
        
\caption{GradCAM++ activation maps for {\mnet} trained on {\gotcha} and tested on {\fows} (top row), and for {\enetPRE} trained on {\fows} and tested on {\gotcha} (bottom row).}
\label{fig:gradcam}
\end{figure}

%In turn, both {\wocc} and {\woocc} might contain shared discriminative statistical cues between genuine and manipulated data
%Thus, in the hypothesis that models trained on {\wocc} actually leverage those artifacts, their performance on the {\woocc} should decrease significantly as 

%\subsection{Swapping Algorithm Mismatch}
%{\it Ongoing}

%\subsection{Occlusion Challenge Mismatch}
%{\it Ongoing}

\section{Discussions and Conclusions}

We have performed an experimental analysis exploring the effectiveness of challenge-based detection for face swapping in video communication scenarios \cite{gotcha}. In those systems, the presence of visual artifacts supposedly introduced by swapping algorithms when dealing with face occlusions is leveraged. Although it is clear how this can substantially help human-based detection, it is conjectured in \cite{gotcha} that they can also improve automated detectors in separating original and manipulated videos, although providing evidence for a single dataset.
We therefore developed the {\fows} dataset, our own video data corpus containing face occlusions in genuine and manipulated videos, so as to extend the empirical evaluation of learning-based automated detectors in a multi-source setting. We trained different baseline and state-of-the-art detectors on both {\fows} and {\gotcha} datasets and performed a comparative analysis. 
% removed: thus, [...] of video manipulation
The prominent finding is an overall lack of model generalization in the cross-dataset setting for all networks, thus confirming a known concern for automated detectors of video manipulation \cite{age-synthetic-reality}\cite{media-forensics-and-df}. In fact, models achieve excellent performance when trained and tested on the same dataset (even with limited training data as in {\fows}), thus effectively picking up intra-dataset discriminative cues between classes. However, it is shown that those do not transfer in the inter-dataset mode, despite some cases where the output scores of CNNs are fairly separated but strongly polarized towards zero. 
In fact, by studying performance when mixing data that contain and do not contain occlusion-based visual artifacts, it emerges that the models learn mostly dataset-specific statistical cues rather than actual occlusion-related cues. 

In other words, our analysis reveals that in order to reliably exploit the advantage given by challenge-based visual artifacts, specialized automated approaches are needed. Assessing strategies for cross-dataset generalization and a broader pool of models proposed in the literature will be the subject of future work.
Also, techniques for automatically guiding the networks towards the relevant face areas will be explored (such as the addition of an occlusion detection module), so that the detector can be activated only when an occlusion is happening and where related artifacts are located. Moreover, one-class detectors and anomaly detection frameworks also represent a promising direction, as they would reduce the dependency on the swapping algorithms used in training.

\section*{Acknowledgment}

We thank the authors of \cite{gotcha} for sharing with us their data before the official release.

This work was partially supported by project SERICS (PE00000014) under the MUR National Recovery and Resilience Plan funded by the European Union - NextGenerationEU.

% \todo{aggiungere ringraziamenti Serics. Forse Metafora o IPZS?}
% \todo{chiedere Silvio/Roberto}

% \todo{sistemare references: arxiv (controllare che siano pubblicati)}
% \Note{
% \begin{itemize}
%     \item mettere solo iniziale nome e congome per intero degli autori
%     \item conferenze: mettere solo autore, titolo, venue, anno. Senza "2017 workshop ecc". Es. {G. Mittal, C. Hegde, and N. Memon. Gotcha: Real-time
% video deepfake detection via challenge-response. In IEEE 
% European Symposium on Security and Privacy (EuroS\&P),
% 2024}
%     \item journal: tenere volume e numero di pagine
%     \item arxiv cercare versione pubblicata e sistemare. Altrimenti viene fuori solo il titolo.
%     \item for github repo : mandelli2020modified
% \end{itemize}
% }

\bibliographystyle{plain} 
\bibliography{Bibliography_noDOI}

\begin{thebibliography}{10}

\bibitem{df2024}
I.~Amerini, M.~Barni, S.~Battiato, P.~Bestagini, G.~Boato, T.~Sari Bonaventura, V.~Bruni, R.~Caldelli, F.~De Natale, R.~De Nicola, L.~Guarnera, S.~Mandelli, G.~L. Marcialis, M.~Micheletto, A.~Montibeller, G.~Orru', A.~Ortis, P.~Perazzo, G.~Puglisi, D.~Salvi, S.~Tubaro, C.~Melis Tonti, M.~Villari, and D.~Vitulano.
\newblock Deepfake media forensics: State of the art and challenges ahead.
\newblock {\em arXiv:2408.00388}, 2024.

\bibitem{MilanPaper}
N.~Bonettini, E.~D. Cannas, S.~Mandelli, L.~Bondi, P.~Bestagini, and S.~Tubaro.
\newblock Video face manipulation detection through ensemble of {CNN}s.
\newblock In {\em International Conference on Pattern Recognition (ICPR)}, 2021.
\newblock Available at \url{https://github.com/polimi-ispl/icpr2020dfdc/tree/master}.

\bibitem{SimSwap_ChenCNG20}
Renwang C., Xuanhong C., Bingbing N., and Yanhao G.
\newblock {SimSwap}: An efficient framework for high fidelity face swapping.
\newblock In {\em {ACM} International Conference on Multimedia}, 2020.

\bibitem{age-synthetic-reality}
J.~P. Cardenuto, J.~Yang, R.~Padilha, R.~Wan, D.~Moreira, H.~Li, S.~Wang, F.~Andaló, S.~Marcel, and A.~Rocha.
\newblock The age of synthetic realities: Challenges and opportunities.
\newblock {\em APSIPA Transactions on Signal and Information Processing}, 12(1), 2023.

\bibitem{xception}
F.~Chollet.
\newblock { Xception: Deep Learning with Depthwise Separable Convolutions }.
\newblock In {\em IEEE Conference on Computer Vision and Pattern Recognition (CVPR)}, 2017.

\bibitem{ENISA_RemoteIDProofing}
{ENISA}.
\newblock {Remote ID Proofing}.
\newblock {ENISA Report}, March 2021.

\bibitem{ENISA_RemoteIDProofingA&C}
{ENISA}.
\newblock {Remote Identity Proofing -- Attacks \& Countermeasures}.
\newblock {ENISA Report}, January 2022.

\bibitem{farid}
C.~R. Gerstner and H.~Farid.
\newblock Detecting real-time deep-fake videos using active illumination.
\newblock In {\em IEEE/CVF Conference on Computer Vision and Pattern Recognition (CVPR) Workshops}, 2022.

\bibitem{pytorchcam}
Jacob Gildenblat and contributors.
\newblock Pytorch library for cam methods.
\newblock 2021.
\newblock Available at \url{https://github.com/jacobgil/pytorch-grad-cam}.

\bibitem{mediapipe-face-detector}
Google.
\newblock Mediapipe face detector python.
\newblock Available at \url{https://developers.google.com/mediapipe/solutions/vision/face\_detector}.

\bibitem{GHOST_Groshev}
A.~Groshev, A.~Maltseva, D.~Chesakov, A.~Kuznetsov, and D.~Dimitrov.
\newblock {GHOST}—a new face swap approach for image and video domains.
\newblock {\em IEEE Access}, 2022.

\bibitem{cr}
E.~Haasnoot, Luuk~J. Spreeuwers, and Raymond N.~J. Veldhuis.
\newblock Presentation attack detection and biometric recognition in a challenge-response formalism.
\newblock {\em EURASIP Journal on Information Security}, (1):5, 2022.

\bibitem{SeeingIsLiving}
C.~Li, L.~Wang, S.~Ji, X.~Zhang, Z.~Xi, S.~Guo, and T.~Wang.
\newblock {Seeing is Living? Rethinking the Security of Facial Liveness Verification in the Deepfake Era}.
\newblock In {\em USENIX Security 22}, pages 2673--2690, 2022.

\bibitem{focal-loss}
T.~Lin, P.~Goyal, R.~Girshick, K.~He, and P.~Dollár.
\newblock Focal loss for dense object detection.
\newblock {\em IEEE Transactions on Pattern Analysis and Machine Intelligence}, 42(2):318--327, 2020.

\bibitem{loshchilov2018decoupled}
I.~Loshchilov and F.~Hutter.
\newblock Decoupled weight decay regularization.
\newblock In {\em International Conference on Learning Representations}, 2019.

\bibitem{gotcha}
G.~Mittal, C.~Hegde, and N.~Memon.
\newblock Gotcha: Real-time video deepfake detection via challenge-response.
\newblock In {\em IEEE European Symposium on Security and Privacy (EuroS\&P)}, 2024.
\newblock Available at \url{https://github.com/mittalgovind/GOTCHA-Deepfakes}.

\bibitem{FSGAN}
Yuval N., Yosi K., and Tal H.
\newblock {FSGAN}: Subject agnostic face swapping and reenactment.
\newblock {\em IEEE/CVF International Conference on Computer Vision (ICCV)}, 2019.

\bibitem{DBLP:conf/pkdd/PasquiniPSR23}
C.~Pasquini, M.~Pernpruner, G.~Sciarretta, and S.~Ranise.
\newblock Towards a fine-grained threat model for video-based remote identity proofing.
\newblock In {\em Machine Learning and Principles and Practice of Knowledge Discovery in Databases - International Workshops of {ECML} {PKDD}}, 2023.

\bibitem{DFL}
I.~Perov, D.~Gao, N.~Chervoniy, K.~Liu, S.~Marangonda, C.~Umé, Dpfks, C.~Shift Facenheim, L.~RP, J.~Jiang, S.~Zhang, P.~Wu, B.~Zhou, and W.~Zhang.
\newblock {DeepFaceLab}: Integrated, flexible and extensible face-swapping framework.
\newblock {\em arXiv: 2005.05535}, 2021.

\bibitem{chapter2022}
M.~M. Pic, G.~Mahfoudi, A.~Trabelsi, and J.~Dugelay.
\newblock {\em Face Manipulation Detection in Remote Operational Systems}.
\newblock Springer International Publishing, 2022.

\bibitem{pres2017}
R.~Ramachandra and C.~Busch.
\newblock Presentation attack detection methods for face recognition systems: A comprehensive survey.
\newblock {\em ACM Computing Surveys}, 50(1), 2017.

\bibitem{Facedancer_Rosberg}
F.~Rosberg, E.~E. Aksoy, F.~Alonso-Fernandez, and C.~Englund.
\newblock {FaceDancer}: Pose- and occlusion-aware high fidelity face swapping.
\newblock In {\em IEEE/CVF Winter Conference on Applications of Computer Vision (WACV)}, 2023.

\bibitem{ff++}
A.~R\"ossler, D.~Cozzolino, L.~Verdoliva, C.~Riess, J.~Thies, and M.~Nie{\ss}ner.
\newblock Faceforensics++: Learning to detect manipulated facial images.
\newblock In {\em ICCV}, 2019.

\bibitem{bbox_large}
Xianyun S., Beibei D., Caiyong W., Bo~P., and Jing D.
\newblock Visual realism assessment for face-swap videos.
\newblock {\em arXiv: 2302.00918}, 2023.

\bibitem{face-manipulation-and-fake-detection}
R.~Tolosana, R.~Vera-Rodr{\'i}guez, J.~Fierrez, A.~Morales, and J.~Ortega-Garcia.
\newblock Deepfakes and beyond: A survey of face manipulation and fake detection.
\newblock {\em arXiv: 2001.00179}, 2020.

\bibitem{stamm}
Danial~Samadi Vahdati, Tai~Duc Nguyen, and Matthew~C. Stamm.
\newblock Defending low-bandwidth talking head videoconferencing systems from real-time puppeteering attacks.
\newblock In {\em IEEE/CVF Conference on Computer Vision and Pattern Recognition (CVPR) Workshops}, 2023.

\bibitem{media-forensics-and-df}
L.~Verdoliva.
\newblock Media forensics and deepfakes: An overview.
\newblock {\em IEEE Journal of Selected Topics in Signal Processing}, 14(5):910–932, 2020.

\bibitem{DFB}
Z.~Yan, Y.~Zhang, Xi. Yuan, S.~Lyu, and B.~Wu.
\newblock Deepfakebench: A comprehensive benchmark of deepfake detection.
\newblock In {\em Advances in Neural Information Processing Systems}, volume~36, pages 4534--4565, 2023.
\newblock Available at \url{https://github.com/SCLBD/DeepfakeBench}.

\end{thebibliography}

\end{document}